\documentclass[11pt,a4paper]{article}
\usepackage{times,latexsym}
\usepackage{url}
\usepackage[normalem]{ulem}
\usepackage[T1]{fontenc}

\usepackage[acceptedWithA]{tacl2018v2}

\usepackage{xspace,mfirstuc,tabulary}

\newif\iftaclinstructions
\taclinstructionsfalse %
\iftaclinstructions
\renewcommand{\confidential}{}
\renewcommand{\anonsubtext}{(No author info supplied here, for consistency with
TACL-submission anonymization requirements)}
\newcommand{\instr}
\fi

\iftaclpubformat %

\else

\fi

\usepackage{microtype}
\usepackage{amsmath}
\usepackage{amssymb}
\usepackage{tikz,pgfplots,pgfplotstable}
\usepackage[inline]{enumitem}
\usetikzlibrary{positioning,shapes,shadows,arrows,automata,chains,fit}
\usepackage{multirow}
\usepackage{rotating}
\usepackage{mathtools}
\newcommand{\lform}[1]{\textsf{\scriptsize{#1}}}
\newcommand{\xml}[1]{$<${\fontfamily{cmtt}\selectfont {#1}}$>$}
\newcommand{\cmtt}[1]{{\fontfamily{cmtt}\selectfont {#1}}}

\DeclarePairedDelimiterX{\infdivx}[2]{(}{)}{%
  #1\;\delimsize\|\;#2%
}
\newcommand{\infdiv}{D\infdivx}

\makeatletter
\newcommand{\thickhline}{%
    \noalign {\ifnum 0=`}\fi \hrule height 1pt
    \futurelet \reserved@a \@xhline
}
\makeatother

\title{Data-to-text Generation with Variational Sequential Planning}

\author{Ratish Puduppully  \textnormal{and} Yao Fu \textnormal{and} Mirella Lapata\\
Institute for Language, Cognition and Computation\\
School of Informatics, University of Edinburgh\\
 10 Crichton Street, Edinburgh EH8 9AB\\
\texttt{r.puduppully@sms.ed.ac.uk}~~~~\texttt{yao.fu@ed.ac.uk}~~~~\texttt{mlap@inf.ed.ac.uk}\\
}

\date{}

\begin{document}
\maketitle
\begin{abstract}
  We consider the task of data-to-text generation, which aims to create
  textual output from non-linguistic input. We focus on generating
  long-form text, i.e.,~documents with multiple paragraphs, and
  propose a neural model enhanced with a planning component
  responsible for organizing high-level information in a coherent and
  meaningful way.  We infer \emph{latent} plans sequentially with a
  structured variational model, while interleaving the steps of
  planning and generation. Text is generated by conditioning on
  previous variational decisions \emph{and} previously generated
  text. Experiments on two data-to-text benchmarks (\textsc{RotoWire}
  and MLB) show that our model outperforms strong baselines and is
  sample efficient in the face of limited training data (e.g., a few
  hundred instances).
\end{abstract}

\begin{figure*}[t!]
\begin{minipage}{0.4\textwidth}
\footnotesize
\begin{tabular}{@{}l@{~~}r@{~~}c@{~~}r@{~~}c@{~~}c@{~~}c@{~~}c@{~~}c@{~~}l@{}} 
\multicolumn{10}{@{}l@{}}{(A)} \\ 
\hline
\lform{TEAM}      & \lform{Inn1} &\lform{Inn2} &\lform{Inn3} & \lform{Inn4} & \lform{$\dots$} & \lform{TR}& \lform{TH} & \lform{E} & \lform{$\dots$} \\ \hline
\lform{Orioles} &\lform{1} &\lform{0} &\lform{0} & \lform{0} & \lform{$\dots$} & \lform{2} & \lform{4} & \lform{0} & \lform{$\dots$} \\ 
\lform{Royals} &\lform{1} &\lform{0} &\lform{0} & \lform{3} &\lform{$\dots$} & \lform{9} & \lform{14} & \lform{1} & \lform{$\dots$} \\ \hline
\end{tabular}

\vspace{.3cm}
\footnotesize
\begin{tabular}{@{}l@{~~}c@{~~}c@{~}r@{~~}c@{~~}c@{~~}l@{~~}l@{}} 
\hline
\lform{BATTER} & \lform{H/V} & \lform{AB} & \lform{BR} & \lform{BH}& \lform{RBI} &  \lform{TEAM} & \lform{$\dots$}\\ \hline
\lform{C.Mullins} & \lform{H} & \lform{4} & \lform{2} & \lform{2}& \lform{1} &   \lform{Orioles} & \lform{$\dots$}\\
\lform{J.Villar} & \lform{H} & \lform{4} & \lform{0} & \lform{0}& \lform{0} & \lform{Orioles} & \lform{$\dots$}\\
\lform{W.Merrifield} & \lform{V}& \lform{2} & \lform{3} & \lform{2} & \lform{1}& \lform{Royals} & \lform{$\dots$}\\
\lform{R.O'Hearn} & \lform{V}& \lform{5} & \lform{1} & \lform{3} & \lform{4}&  \lform{Royals} & \lform{$\dots$}\\

\lform{$\dots$} & \lform{$\dots$} & \lform{$\dots$} & \lform{$\dots$} &  \lform{$\dots$} &  \lform{$\dots$} & \lform{$\dots$}\\\hline
\end{tabular}

\vspace{.3cm}
\footnotesize
\begin{tabular}{@{}l@{~~}c@{~~}c@{~}r@{~~}c@{~~}c@{~~}c@{~~}c@{~~}c@{~~}l@{~~}l@{}} \hline
\lform{PITCHER} & \lform{H/V} & \lform{W} & \lform{L} & \lform{IP}& \lform{PH} &  \lform{PR} &  \lform{ER}&  \lform{BB}&  \lform{K} & \lform{$\dots$}\\ \hline
\lform{A.Cashner} & \lform{H} & \lform{4} & \lform{13} & \lform{5.1}& \lform{9} &   \lform{4}&   \lform{4} &   \lform{3}&   \lform{1}& \lform{$\dots$}\\
\lform{B.Keller} & \lform{V}& \lform{7} & \lform{5} & \lform{8.0} & \lform{4}&  \lform{2}  &   \lform{2}&   \lform{2}&   \lform{4} & \lform{$\dots$}\\
\lform{$\dots$} & \lform{$\dots$} & \lform{$\dots$} & \lform{$\dots$} &  \lform{$\dots$} &  \lform{$\dots$} & \lform{$\dots$}\\\hline
\end{tabular}

\vspace{.3cm}                        

\lform{Inn1:} runs in innings, \lform{TR:} team runs, \lform{TH:}  team hits, \lform{E:}  errors,
 \lform{H/V:} home/visiting,
 \lform{AB:} at-bats, \lform{BR:} batter runs, \lform{BH:} batter hits, \lform{RBI:} runs-batted-in,
  \lform{W:} wins, \lform{L:} losses,  \lform{IP:} innings pitched, \lform{PH:} hits given, 
\lform{PR:} runs given, \lform{ER:} earned runs,
  \lform{BB:} walks, \lform{K:} strike outs, \lform{INN:} inning with (T)op/(B)ottom,
\lform{PL-ID:} play id, \lform{SCR}: score of Royals.

\end{minipage}
\begin{minipage}{0.5\textwidth}
\vspace*{-2.5cm}
\footnotesize
\hspace*{-1.2cm}\begin{tabular}{@{}p{9cm}@{}} 
\multicolumn{1}{@{}l@{}}{(C)} \\
\hline 
  
                  {\scriptsize
                  KANSAS CITY, Mo. -- \textcolor{olive}{Brad Keller kept up his recent
                  pitching surge with another strong
                  outing.} \textcolor{blue}{\xml{P}}
		  \textcolor{cyan}{
                  Keller gave up a home run to the first batter of the game -- Cedric Mullins -- but quickly settled in to pitch eight strong innings in the Kansas City Royals' 9--2 win over the Baltimore Orioles in a matchup of the teams with the worst records in the majors.} \textcolor{blue}{\xml{P}}
                  \textcolor{magenta}{Keller (7--5) gave up two runs and four hits with two walks and four strikeouts to improve to 3--0 with a 2.16 ERA in his last four starts}. \textcolor{blue}{\xml{P}}
                  \textcolor{teal}{Ryan O'Hearn homered among his three hits and drove in four runs, Whit Merrifield scored three runs, and Hunter Dozier and Cam Gallagher also went deep to help the Royals win for the fifth time in six games on their current homestand.} \textcolor{blue}{\xml{P}}
		  \textcolor{red}{
                  With the score tied 1--1 in the fourth, Andrew Cashner (4--13) gave up a sacrifice fly to Merrifield after loading the bases on two walks and a single. Dozier led off the fifth inning with a 423-foot home run to left field to make it 3-1.} \textcolor{blue}{\xml{P}}
                  \textcolor{orange}{The Orioles pulled within a run in the sixth when Mullins led off with a double just beyond the reach of Dozier at third, advanced to third on a fly ball and scored on Trey Mancini's sacrifice fly to the wall in right.} \textcolor{blue}{\xml{P}} \lform{$\dots$}
}\\ \hline
\end{tabular}
\end{minipage}
\begin{minipage}{\textwidth}
\vspace{-2.2cm}

\hspace*{6.7cm}
\footnotesize
\begin{tabular}{@{}l@{~}c@{~}c@{~}c@{~}c@{~}r@{~}c@{~}c@{~}l@{}} 
\multicolumn{9}{@{}l@{}}{(B)} \\
\hline
\lform{BATTER}&\lform{PITCHER}&\lform{SCORER}&\lform{ACTION}&\lform{TEAM} & \lform{INN} & \lform{PL-ID} & \lform{SCR} & \lform{$\dots$}\\ \hline
\lform{C.Mullins}&\lform{B.Keller}&\lform{---}&\lform{Home run}&\lform{Orioles} & \lform{1-T} &  \lform{1} &  \lform{1} & \lform{$\dots$}\\ 
\lform{H.Dozier}&\lform{A.Cashner}&\lform{W.Merrifield}&\lform{Grounded}&\lform{Royals} &  \lform{1-B} &\lform{3} &  \lform{1} & \lform{$\dots$}\\ 
\lform{W.Merrifield}&\lform{A.Cashner}&\lform{B.Goodwin}&\lform{Sac fly}&\lform{Royals} & \lform{4-B} &  \lform{5}& \lform{2} & \lform{$\dots$}\\ 
\lform{H.Dozier}&\lform{A.Cashner}&\lform{---}&\lform{Home run}&\lform{Royals} & \lform{5-B}&  \lform{1} & \lform{3} & \lform{$\dots$}\\ 
\lform{$\dots$} & \lform{$\dots$} & \lform{$\dots$} & \lform{$\dots$} &  \lform{$\dots$} & \lform{$\dots$} & \lform{$\dots$} & \lform{$\dots$}\ & \lform{$\dots$}\\\hline
\end{tabular}
\end{minipage}

\raisebox{.83cm}[0pt]{(D)}
\begin{minipage}{\textwidth}
\begin{center}
\footnotesize
\fontfamily{cmtt}\selectfont\footnotesize
 \begin{tabular}{p{0.32\columnwidth}} \\\hline 
    V(Orioles),    V(Royals),\\
   V(C.Mullins),
    V(J.Villar), 
    V(W.Merrifield),
    V(R.O'Hearn),
    V(A.Cashner), 
    V(B.Keller), 
    V(H.Dozier),
    $\dots$, \\ 
V(1-T), 
    V(1-B), 
    V(2-T),
    V(2-B), 
    V(3-T), 
    V(3-B),
$\dots$ \\ \hline
  \end{tabular}
\quad
  \begin{tabular}{p{0.5\columnwidth}} \\\hline 
    V(Royals) V(Orioles), \\
V(Orioles) V(C.Mullins), V(Orioles) V(J.Villar), 
V(Royals) V(W.Merrifield), 
V(Royals) V(R.O'Hearn),  
V(Orioles) V(A.Cashner),
V(Royals) V(B.Keller), $\dots$,\\
V(C.Mullins) V(Royals) V(Orioles), \\
V(J.Villar) V(Royals) V(Orioles),
$\dots$
 \\\hline
  \end{tabular}
  \end{center}
 \end{minipage}

(E)\hspace*{.3cm}\begin{minipage}{\textwidth}
\begin{center}
\footnotesize
  \begin{tabular}{@{}p{14.5cm}@{}} 
   \multicolumn{1}{c}{}\\\hline
\multicolumn{1}{c}{\vspace*{-.2cm}} \\
   {\fontfamily{cmtt}\selectfont\footnotesize

    \textcolor{olive}{V(B.Keller)}\textcolor{blue}{\xml{P}}\textcolor{cyan}{V(B.Keller) V(C.Mullins)
    V(Royals)
    V(Orioles)}\textcolor{blue}{\xml{P}}\textcolor{magenta}{V(B.Keller)}\textcolor{blue}{\xml{P}}  \textcolor{teal}{V(R.O'Hearn)
    V(W.Merrifield) V(H.Dozier) V(C.Gallagher)} \textcolor{blue}{\xml{P}}\textcolor{red}{V(4-B, 5-B)} \textcolor{blue}{\xml{P}}
    \textcolor{orange}{V(6-T)}\textcolor{blue}{\xml{P}}}\\ \hline
  \end{tabular}
  \end{center}
   \end{minipage}
   \caption{Example from the MLB dataset reproduced from
     \citet{Puduppully-2020} with the authors' permission.  Table~A is
     typically referred to as \emph{box score}. It summarizes the data
     of the game per team and player.  Table~B reports statistics
     pertaining to innings or play-by-play scores.  Table~C contains
     the game summary.  Paragraphs in Table C are separated with
     \textcolor{blue}{\xml{P}} delimiters. Table~D contains paragraph
     plans obtained from Tables~A and~B.  Paragraph plans in the
     first column correspond to a single entity or event. Paragraph
     plans in the second column describe combinations of entities or
     events.  \xml{V(entity)} verbalizes records pertaining to
     entities and \xml{V(inning-T/B)} verbalizes records for the
     Top/Bottom side of an inning.  Paragraph plans correspond to 
     paragraphs in Table~C. Table~E contains the \emph{macro plan}
     for the document in Table~C. A macro plan is a sequence of
     paragraph plans.  Plan-document correspondences are highlighted
     using the same color.}
\label{fig:example}
\end{figure*}

\begin{figure}[t]
  \centering
  \includegraphics[width=0.9\linewidth]{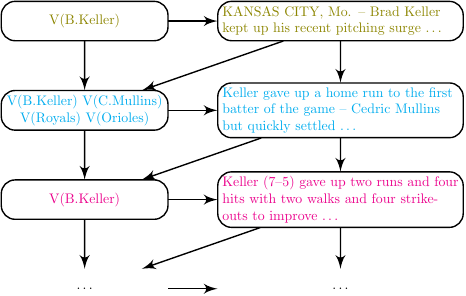}
\vspace*{-1.5ex}
  \caption{Conceptual sequence of interleaved planning and generation steps. The
paragraph plan and its corresponding paragraph have the same color.}
  \label{fig:flow-diag}
\end{figure}

\section{Introduction}
Data-to-text generation refers to the task of generating textual
output from non-linguistic input such as database tables,
spreadsheets, or simulations of physical systems
\cite{Reiter:1997:BAN:974487.974490,reiter-dale:00,DBLP:journals/jair/GattK18}.
Recent progress in this area \cite{mei-etal-2016-talk,
  lebret-etal-2016-neural,wiseman-etal-2017-challenges} has been
greatly facilitated by the very successful encoder-decoder neural
architecture \cite{sutskever2014sequence} and the development of large
scale datasets. \textsc{RotoWire} \cite{wiseman-etal-2017-challenges}
and MLB \cite{puduppully-etal-2019-data} constitute such
examples. They both focus on the sports domain which has historically
drawn attention in the generation community
\cite{barzilay-lapata-2005-collective,tanaka-ishii-etal-1998-reactive,robin1994revision}
and consider the problem of generating long target texts from database
records.

Figure~\ref{fig:example} (reproduced from \citealp{Puduppully-2020})
provides a sample from the MLB dataset which pairs human written
summaries (Table C) with major league baseball game statistics.  These
are mostly scores (collectively referred to as \emph{box score}) which
summarize the performance of teams and players, e.g., batters,
pitchers, or fielders (Table~A) and a \emph{play-by-play} description
of the most important events in the game (Table~B). Game summaries in
MLB are relatively long (540 tokens on average) with multiple
paragraphs (15~on average). %
The complexity of the input and the length of the game
summaries pose various challenges to neural models which, despite
producing fluent output, are often imprecise, prone to hallucinations,
and display poor content selection
\cite{wiseman-etal-2017-challenges}.  Attempts to address these issues
have seen the development of special-purpose modules which keep track
of salient entities
\cite{iso-etal-2019-learning,puduppully-etal-2019-data}, determine
which records (see the rows in Tables~A and~B) should be mentioned in
a sentence and in which order
\cite{DBLP:journals/corr/abs-1809-00582,narayan-etal-2020-stepwise},
and reconceptualize the input in terms of paragraph plans
\cite{Puduppully-2020} to facilitate document-level planning (see
Table~D in Figure~\ref{fig:example}).

Specifically, \citet{Puduppully-2020} advocate the use of \emph{macro
  plans} for improving the organization of document content and
structure. A {macro plan} is a \emph{sequence} of paragraph plans, and
each paragraph plan corresponds to a document paragraph. A macro plan
is shown in Table~E (Figure~\ref{fig:example}).  Examples of paragraph
plans are given in Table~D where \xml{V(entity)} verbalizes records
pertaining to entities and \xml{V(inning-T/B)} verbalizes records for
the Top/Bottom side of an inning.  Verbalizations are sequences of
record types followed by their values. Document paragraphs are shown
in Table~C and have the same color as their corresponding plans in
Table~E.  During training, \citet{Puduppully-2020} {\it learn} to
predict a macro plan from a pool of paragraph plans, and produce a
game summary based on it. Continuing with our example in
Figure~\ref{fig:example}, plan~(E) is obtained from paragraph
plans~(D), to give rise to game summary~(C).

The intermediate macro plan renders generation more interpretable
(differences in the output can be explained by differences in macro
planning). It also makes modeling easier, the input is no longer a
complicated table but a sequence of paragraph plans which in turn
allows us to treat data-to-text generation as a sequence-to-sequence
learning problem. Nevertheless, decoding to a long document remains
challenging for at least two reasons. Firstly, the macro plan may be
encoded as a sequence but a very long one (more than 3,000 tokens)
which the decoder has to attend to at each time step in order to
generate a summary token-by-token. Secondly, the prediction of the
macro plan is conditioned solely on the input (i.e.,~pool of paragraph
plans~(D) in Figure~\ref{fig:example}) and does not make use of
information present in the summaries. We hypothesize that planning
would be more accurate were it to consider information available in the
table (and corresponding paragraph plans) and the generated summary,
more so because the plans are coarse-grained and there is a
one-to-many relationship between a paragraph plan and its realization.
For example, we can see that the plan for \xml{V(B.Keller)} results in
two very different realizations in the summary in
Figure~\ref{fig:example} (see first and third paragraph).

In this work, we present a model which interleaves macro planning with
text generation (see Figure \ref{fig:flow-diag} for a sketch of the
approach).  We begin by selecting a plan from a pool of paragraph
plans (see Table~D in Figure~\ref{fig:example}), and generate the
first paragraph by conditioning on it.  We select the next plan by
conditioning on the previous plan \emph{and} the previously generated
paragraph. We generate the next paragraph by conditioning on the
currently selected plan, the previously predicted plan, and generated
paragraph.  We repeat this process until the final paragraph plan is
predicted.  We model the selection of paragraph plans as a
\emph{sequential latent variable process} which we argue is intuitive
since content planing is inherently latent. Contrary to
\citet{Puduppully-2020}, we do not a priori decide on a \emph{global}
macro plan. Rather our planning process is \emph{incremental} and as a result
less rigid. Planning is informed by generation and vice versa, which
we argue should be mutually beneficial (they are conditioned on each
other).

During training, the sequential latent model can better leverage the
summary to render paragraph plan selection more accurate and take
previous decisions into account.  We hypothesize that the
interdependence between planning and generation allows the model to
cope with diversity.  In general, there can be many ways in which the
input table can be described in the output summary, i.e., different
plans give rise to equally valid game summaries. The summary in
Figure~\ref{fig:example} (Table C) focuses on the performance of
\textsl{Brad Keller} who is a high scoring pitcher (first three
paragraphs). An equally plausible summary might have discussed a high
scoring batter first (e.g., \textsl{Ryan O'Hearn}). Also notice that
the summary describes innings in chronological order.  However,
another ordering might have been equally plausible, for example,
describing innings where the highest runs are scored first or innings
which are important in flipping the outcome of the match. In the face
of such diversity, there may never be enough data to learn an accurate
global plan. It is easier to select a paragraph plan from the pool
once some of the summary is known, and different plans can be
predicted for the same input. 
In addition, the proposed 
model is end-to-end differentiable and gradients for summary 
prediction also inform plan prediction.

Our contributions can be summarized as follows: (1) we decompose data-to-text generation into sequential plan
      selection and paragraph generation. The two processes are
      interleaved and generation proceeds incrementally. We look at what has been already generated,
      make a plan on what to discuss next, realize the plan, and
      repeat; (2) in contrast to previous
      models \cite{DBLP:journals/corr/abs-1809-00582,Puduppully-2020}
      where content plans are monolithic and determined in advance,
      our approach is more flexible, it simplifies modeling (we do not
      need to learn alignments between paragraph plans and summary
      paragraphs), and leads to sample efficiency in
      low resource scenarios; 
      (3)~our approach scales better for tasks involving generation of
      long multi-paragraph texts, as we do not need to specify the
      document plan in advance; %
      (4)~experimental results on English and German \textsc{RotoWire}
      \cite{wiseman-etal-2017-challenges,hayashi-etal-2019-findings},
      and MLB \cite{puduppully-etal-2019-data} show that our model is
      well-suited to long-form generation and generates more factual,
      coherent, and less repetitive output compared to strong
      baselines. 
      
      We share our code and models in the hope of being
      useful for other tasks (e.g., story generation, summarization)\footnote{\url{https://github.com/ratishsp/data2text-seq-plan-py}}.

\section{Related Work}

A long tradition in natural language generation views content planning
as a central component to identifying important content and
structuring it appropriately \cite{reiter-dale:00}.  Earlier work has
primarily made use of hand-crafted content plans with some exceptions
which pioneered learning-based approaches. For instance,
\citet{duboue-mckeown-2001-empirically} learn ordering constraints on
the content plan, while \citet{kan-mckeown-2002-corpus} learn content
planners from semantically annotated corpora, and
\citet{konstas-lapata-2013-inducing} predict content plans using
grammar rules whose probabilities are learnt from training data. 

More recently, there have been attempts to equip encoder-decoder
models
\cite{DBLP:journals/corr/BahdanauCB14,wiseman-etal-2017-challenges}
with content planning modules.
\citet{DBLP:journals/corr/abs-1809-00582} introduce \emph{micro
  planning}: they first learn a content plan corresponding to a
sequence of records, and then generate a summary conditioned on it.
\citet{narayan-etal-2020-stepwise} treat content selection as a task
similar to extractive summarization. Specifically, they post-process
Pudupully et al.'s \shortcite{DBLP:journals/corr/abs-1809-00582}
micro-plans with special tokens identifying the beginning and end of a
sentence. Their model first extracts sentence plans and then
verbalizes them one-by-one by conditioning on previously generated
sentences.
\citet{moryossef-etal-2019-step,moryossef-etal-2019-improving} propose
a two-stage approach which first predicts a document plan and then
generates text based on it. The input to their model is a set of RDF
$\langle$Subject, Object, Predicate$\rangle$ tuples.  Their document
plan is a sequence of sentence plans where each sentence plan contains
a subset of tuples in a specific order.  Text generation is
implemented using a sequence-to-sequence model enhanced with attention
and copy mechanisms \cite{DBLP:journals/corr/BahdanauCB14}. They
evaluate their model on the WebNLG dataset
\citep{gardent-etal-2017-creating} where the outputs are relatively
short (24 tokens on average).

Our approach is closest to \citet{Puduppully-2020} who advocate
\emph{macro planning} as a way of organizing high-level document
content. Their model operates over paragraph plans which are
verbalizations of the tabular input and predicts a document plan as a
sequence of paragraph plans.  In a second stage, the summary is
generated from the predicted plan making use of attention enriched
with a copy mechanism.  We follow their formulation of content
planning as paragraph plan prediction. Our model thus operates over
larger content units compared to related work
\cite{DBLP:journals/corr/abs-1809-00582,narayan-etal-2020-stepwise}
and performs the tasks of micro- and macro-planning in one go. In
contrast to \citet{Puduppully-2020}, we predict paragraph plans and
their corresponding paragraphs \emph{jointly} in an incremental
fashion. Our approach is reminiscent of psycholinguistic models of
speech production
\cite{levelt1993speaking,taylor-taylor-1990-book,guhe2020incremental}
which postulate that different levels of processing (or modules) are
responsible for language generation; these modules are incremental,
each producing output as soon as the information it needs is available
and the output is processed immediately by the next module.

We assume plans form a sequence of paragraphs which we treat as a
latent variable and learn with a structured variational
model. Sequential latent variables
\cite{NIPS2015_b618c321,NIPS2016_208e43f0,NIPS2017_900c563b} have
previously found application in modeling attention in
sequence-to-sequence networks \cite{shankar2018posterior}, document
summarization \cite{li-etal-2017-deep}, controllable generation
\cite{li-rush-2020-posterior,NEURIPS2020_ea119a40}, and
knowledge-grounded dialogue \cite{Kim2020Sequential}.  In the context
of data-to-text generation, latent variable models have been primarily
used to inject diversity in the output.  \citet{shao-etal-2019-long}
generate a sequence of groups (essentially a subset of the input)
which specifies the content of the sentence to be generated. Their
plans receive no feedback from text generation, they cover a small set
of input items, and give rise to relatively short documents
(approximately 100~tokens long).  \citet{Ye2020Variational} use latent
variables to disentangle the content from the structure
(operationalized as templates) of the output text.  Their approach
generates diverse output output by sampling from the template-specific
sample space.  They apply their model to single-sentence generation
tasks \cite{lebret-etal-2016-neural,reed-etal-2018-neural}.

\section{Model}
\begin{figure*}[t]
  \centering
  \includegraphics[width=0.9\linewidth]{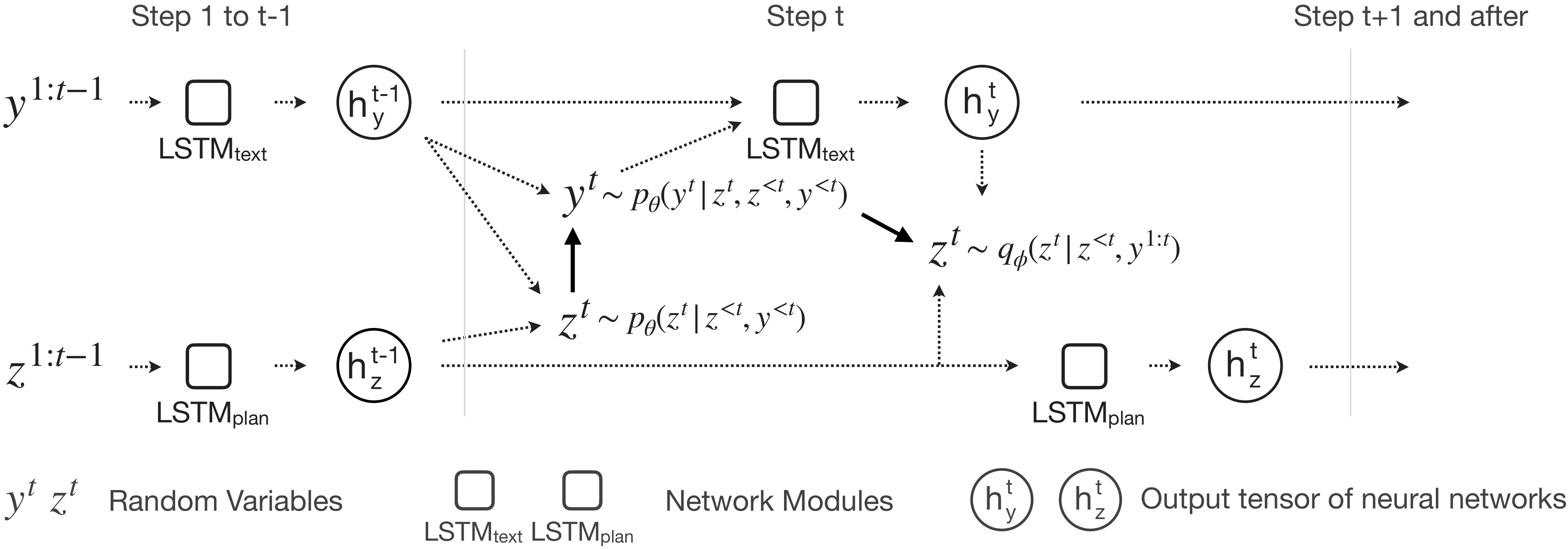}
\vspace*{-1.5ex}
  \caption{ Model workflow.  Solid arrows show dependencies between
    random variables.  Dashed arrows show the computation graph whose
    backbone consists of an LSTM$_{\text{text}}$ and an
    LSTM$_\text{plan}$.  Note that the variational model and the
    generative model are tied closely with the shared LSTM.  To
    generate long documents, the model observes what has been already
    generated, decides on a plan about what to discuss next, uses this
    plan to guide next stage generation, and repeats until the end.  }
  \label{fig:model}
\end{figure*}

Following \citet{Puduppully-2020}, we assume that at training time our
model has access to a pool of paragraph plans $\mathcal{E}$ (see
Table~D in Figure \ref{fig:example}) which represent a clustering of
records. We explain how paragraph plans are created from tabular input
in Section~\ref{sec:experimental-setup}.  Given~$\mathcal{E}$, we aim
to generate a sequence of paragraphs $y = [y^1, ..., y^T]$ that
describe the data following a sequence of chosen plans $z = [z^1, ...,
z^T]$. Let~$y^t$ denote a paragraph, which can consist of multiple
sentences, and $T$~the count of paragraphs in a summary.  With a
slight abuse of notation, superscripts denote indices rather than
exponentiation. So, $y^t_i$~refers to the $i$-th word in the $t$-th
paragraph. A plan $z = [z^1, ..., z^T]$ is a list of discrete
variables where~$z^t = j$ means that we choose the~$j$-th item from
pool~$\mathcal{E}$ of candidate plans to guide the generation of
paragraph~$y^t$.

\paragraph{Generation with Latent Plans} 
The core technique of our model is learning the sequence of latent
plans that guides long document generation. We consider a conditional
generation setting where the input~$\mathcal{E}$ is a set of paragraph
plans and the output $y_{1:T}$ are textual paragraphs verbalizing the
selected sequence $z=z_{1:T}$.  Our goal is to induce variables~$z$
that indicate which paragraphs are being talked about and in which
order.  Similar to previous work
\cite{li-rush-2020-posterior,NEURIPS2020_ea119a40}, we model this
process as a conditional generative model that produces both~$y$
and~$z$ and factorizes as:
\begin{align}
    &p_{\theta}(y, z| \mathcal{E}) = \notag \\
    &\prod_t p_{\theta}(z^t|y^{<t},z^{<t},\mathcal{E}) p_{\theta}(y^t|y^{<t},z^{1:t},\mathcal{E})   \label{eq:p-y-given-k}
\end{align}
where $\theta$ denotes the model parameters and~$<t$ all indices
smaller than~$t$.  We believe this formulation is intuitive,
simulating incremental document generation: inspect~$y^{<t}$ (what has
been already said), make a plan~$z^t$ about what to say next, realize
this plan by generating a new paragraph~$y^t$, and so on.

\paragraph{Inference Model} We are interested in the 
posterior distribution $p_{\theta}(z|y,\mathcal{E})$, i.e.,~the
probability over plan sequences~$z$ for a known text~$y$ and
input~$\mathcal{E}$. This distribution is intractable to compute in
general as the summation of all possible plan sequences~$z$ is
exponentially complex:
\begin{align}
 p_{\theta}(z|y,\mathcal{E}) = \frac{p_\theta(y, z | \mathcal{E})}{\sum_{z} p_\theta(y, z | \mathcal{E})}
\end{align} 

We use variational inference
\cite{DBLP:journals/corr/KingmaW13,DBLP:conf/icml/RezendeMW14} to
approximate the posterior with a parametrized
distribution~$q_{\phi}(z|y,\mathcal{E})$ from which we sample values
of~$z$ that are likely to produce~$y$ (see
\citealt{DBLP:journals/corr/Doersch16} for a tutorial on this
topic). Specifically, we employ an autoregressive inference model
factorized as:
\begin{align}
    q_{\phi}(z|y,\mathcal{E}) &= \prod_t q_{\phi}(z^t|y^{1: t},z^{<t},\mathcal{E}) \label{eq:p-e-given-k}
\end{align}
Note that a major difference between~$q$ above and~$p$ in
Equation~\eqref{eq:p-y-given-k} is that $p$ generates $y_t$ under the
guidance of $z_t$ (conceptually $z^t \to y^t$) while $q$ infers $z_t$
given \emph{observed}~$y_t$ (conceptually $y^t \to z^t$).

\paragraph{Neural Parametrization} 
At step $t$, we start with the encoding of previous
paragraphs~$y^{<t}$ and plans~$z^{<t}$ (see Figure~\ref{fig:model}
left).  Following \citet{yang-etal-2016-hierarchical}, we use a
Bi-directional LSTM (BiLSTM) with a self-attention layer to encode
paragraph~$y^t$ as a vector~$r_y^t$ at step~$t$:
\begin{align}
  r_y^t = \operatorname{Attn}(q_{\text{text}},
  \operatorname{BiLSTM}(y^t)) \label{eq:y_enc}
\end{align}
where $q_{\text{text}}$ is a trainable query vector, which is randomly initialized and learnt along with
the rest of the parameters. 
$\operatorname{Attn}(\cdot)$ returns the attention probability 
and output vector over $\operatorname{BiLSTM}$ representation~$y^t$
with query vector~$q_{\text{text}}$.\footnote{In our notation neural
  network layers are described by math functions.}  Our model uses the
output vector. Next, we encode $r_y^{<t}$ with LSTM$_\text{text}$ as:
\begin{align}
    h_y^{<t} = \operatorname{LSTM}_{\text{text}}(r_y^{<t})
\end{align}
We encode candidate plans in pool \mbox{$\mathcal{E} = [e_{1},
  .., e_{N}]$} with a BiLSTM, similar to the paragraph encoding shown
in Equation~\eqref{eq:y_enc}, and select one of them at each step.
Let~$r_{z}^t$ denote a plan embedding at step~$t$. We
encode~$r_{z}^{<t}$ using LSTM$_\text{plan}$ as:
\begin{align}
    h_{z}^{<t} &= \operatorname{LSTM}_{\text{plan}}(r_{z}^{<t})
\end{align}
The currently selected plan is parametrized as:
\begin{align}
    &h^{t-1} = \operatorname{FF}_{\text{plan}}([h_{z}^{t-1} ; h_y^{t-1}]) \label{eq:h_in_p}\\ 
    & p_\theta(z^t | y^{<t}, z^{<t}, \mathcal{E}) = \operatorname{Attn}(h^{t-1},\mathcal{E}) \label{eq:p_plan}
\end{align}
where $h^{t-1}$ summarizes information in~$y^{<t}$ and~$z^{<t}$,
$\operatorname{FF}_{\text{plan}}(\cdot)$ denotes a feed-forward layer,
and $\operatorname{Attn}(\cdot)$ returns the attention probability
(and output vector) of choosing a plan from $\mathcal{E}$ with current
state $h^{t-1}$. %
Here, we use the attention distribution, which serves essentially as a
copy mechanism.  Then, a plan~$z^t$ is sampled from~$p$ (we use greedy
decoding in our experiments), and its representation~$r_{z}^t$ is used
to update {LSTM}$_{\text{plan}}$ (Figure~\ref{fig:model} right):
\begin{align}
  h_{z}^t = \operatorname{LSTM}_\text{plan}(r_{z}^t, h_{z}^{t-1})
\end{align}

We guide the generation of~$y^t$ with current plan~$z^t$ and decode
each word~$y^t_i$ sequentially with an {LSTM}$_\text{gen}$ decoder
which makes use of beam search. Let~$s_i$ denote the $i$-th decoder
state (initialized with the plan encoding).
We update it as:
\begin{align}
s_{i} = \operatorname{LSTM}_\text{gen}(y^t_{i-1}, s_{i-1}, h^{t-1}_{y})
\end{align}
Note that we feed~$h^{t-1}_{y}$, representing the context of previous
paragraphs, as additional input similar to
\citet{DBLP:conf/aaai/SerbanSLCPCB17}.
Let $r_{z, 1}^t, ..., r_{z, l}^t$ denote the encoding of tokens of the
current plan where $r_{z, k}^t$ is the output of the BiLSTM plan
encoder and $l$ the length of the chosen plan.  We generate the next
word as:
\begin{align}
    &c_{i} = \operatorname{Attn}(s_i, [r_{z, 1}^t, ..., r_{z, l}^t]) \label{eq:cross_attn} \\ 
    &p_\theta(y^t_i | z^t, y^t_{1:i-1}, y^{<t}, z^{<t}, \mathcal{E}) = \notag \\
    &\quad \quad \quad\quad\quad \quad\quad \operatorname{softmax}(\operatorname{FF_{\text{gen}}}([s_i; c_{i}])) \label{eq:generation} 
\end{align}
where $c$ denotes the context vector. 
In Equation \ref{eq:cross_attn}, we use the output vector from $\operatorname{Attn}(\cdot)$. 
$\operatorname{FF}_{\text{gen}}(\cdot)$ 
represents a feed-forward layer. 
In addition, we equip the decoder with copy attention \cite{see-etal-2017-get} to 
enable copying tokens from~$z^t$. As part of this, we learn a probability for 
copy based on $s_i$ \cite{gehrmann-etal-2018-bottom}.
Once paragraph~$y^t$ has been generated, we obtain its
encoding~$r_y^t$ with Equation~\eqref{eq:y_enc}, and update
LSTM$_{\text{text}}$ (Figure~\ref{fig:model} middle):
\begin{align}
    h_y^t = \text{LSTM}_\text{text}(r_y^t, h_y^{t-1})
\end{align}

We parametrize the variational model so that it shares the LSTMs for
encoding~$y$ and $\mathcal{E}$ with the generative model:
\begin{align}
    &\tilde{h}^t = \operatorname{FF_{\text{v}}}([h_{z}^{t-1} ; h_y^t]) \label{eq:h_in_q}\\
    &q_\phi(z^t | y^{1:t}, z^{<t}, \mathcal{E}) = \operatorname{Attn}(\tilde{h}^t, \mathcal{E}) \label{eq:planner}
\end{align}
where $\operatorname{FF}_{\text{v}}(\cdot)$ represents a feed-forward layer. 
Note that Equation~\eqref{eq:h_in_q} differs from
Equation~\eqref{eq:h_in_p} in that it uses the updated~$h^t_y$ instead
of the previous~$h^{t-1}_y$ because now~$y^t$ is observed.  The
variational distribution is again parametrized by the attention
probability.  Essentially, $p$ and $q$ are strongly tied to each other
with the shared LSTM encoders.

Although we primarily focus on the inference, and how the latent plan
can improve the generation of long documents, we note that the model
sketched above could be parametrized differently, e.g., by replacing
the encoder and decoder with pretrained language models like
BART~\citep{lewis-etal-2020-bart}. However, we leave this to future
work.
\paragraph{Training} We optimize the standard evidence lower bound
(ELBO) loss:
\begin{align}
\hspace*{-.2cm}\mathcal{L}_0 = & \log p_{\theta}(y|\mathcal{E}) -
\infdiv{q_{\phi}(z|y,\mathcal{E})}{p_{\theta}(z|y,\mathcal{E})} \notag \\
 &\hspace*{-.4cm}=\mathbb{E}_{q_{\phi}(z| y, \mathcal{E})}[\log p_{\theta}(y, z|\mathcal{E}) - \log q_{\phi}(z | y, \mathcal{E})] \notag \\ %
  =& \mathbb{E}_{q_{\phi}(z| y, \mathcal{E})}\Big[\sum_t \Big\{\log p_{\theta}(y^t | z^t, y^{<t}, z^{<t}, \mathcal{E}) + \notag\\
    & \log \big(\frac{p_{\theta}(z^t | y^{<t}, z^{<t}, \mathcal{E})}{q_{\phi}(z^t | y^{1:t}, z^{<t}, \mathcal{E})}\Big)\Big\}\Big]
\end{align}
where~$\log p_{\theta}(y|\mathcal{E})$ %
is the log-evidence from the data, and
$\infdiv{q_{\phi}(z|y,\mathcal{E})}{p_{\theta}(z|y,\mathcal{E})}$ is the
Kullback-Leibler divergence between~$q_{\phi}$ and the true posterior~$p_{\theta}$.
The objective eventually decomposes to a summation of the reconstruction
probability $p_\theta(y^t|\cdot)$ and the ratio between
$p_\theta(z^t|\cdot)$ and $q_\phi(z^t|\cdot)$ at each step.

Advantageously, we can exploit oracle plans (see Table E in
Figure~\ref{fig:example} and the description in
Section~\ref{sec:experimental-setup} for how these were created) to
obtain weak labels~$z^*$ which we use as distant supervision to the
inference model:
\begin{align}
    \mathcal{L}_{1} &= \mathbb{E}_{z^*}[\log q_{\phi}(z^* | y, \mathcal{E})] \label{eq:training_supervision} \\ 
    \mathcal{L} &= \mathcal{L}_{0} + \lambda \mathcal{L}_{1} \label{eq:total_loss}
\end{align}
Such distant supervision is essential for stabilizing training (it
would be extremely challenging to optimize the model in a fully
unsupervised way) and for mitigating posterior collapse. We use
Gumbel-Softmax~\citep{maddison2016concrete,jang2017categorical} for
differentiable sampling (reparameterization) from~$q$.  The model is
trained with scheduled sampling \cite{10.5555/2969239.2969370}, and
follows the curriculum learning strategy using linear decay
scheduling.  During earlier stages of training predicted plans are
less accurate, and we thus sample from oracle plans at a rate which
decays linearly with training:
\begin{align}
    \epsilon_k &= \max(0, 1 - c*k) \label{eq:scheduled_sampling}
\end{align}
where $c$ is the slope of the decay at training step~$k$.

\section{Experimental Setup}
\label{sec:experimental-setup}

\begin{table}[t]
  \footnotesize
\centering
  \begin{tabular}{@{}p{3cm}ccc@{}} \thickhline 
  & RW & MLB & DE-RW\\ 
\thickhline 
Vocab Size & 11.3K & 38.9K  & 9.5K \\ 
\# Tokens & 1.5M & 14.3M & 234K\\ 
\# Instances & 4.9K & 26.3K & 723 \\ 
\# Paragraphs & 399K & 47.7K & 7K \\
\# Record Types & {39} & 53 & 39 \\ 
Avg Records & 628 & 565 & 628 \\ 
Avg  Length & 337.1 & 542.1 & 323.6 \\ 
Avg Plan length & 10.6  & 15.1  & 9.5 \\ 
\thickhline  
\end{tabular}
\caption{\label{dataset-stats}
Dataset statistics for  \textsc{RotoWire} (RW), MLB and German \textsc{RotoWire}
(DE-RW).  Vocabulary size,
number of tokens, number of instances (i.e., table-summary pairs),
number of paragraphs,
number of record types, average number of records, %
average summary length, average 
macro plan length measured in terms of number of paragraphs.}
    \end{table}

\paragraph{Data}
We performed experiments on the \textsc{RotoWire}
\cite{wiseman-etal-2017-challenges} and MLB
\cite{puduppully-etal-2019-data} datasets and the German
\textsc{RotoWire} provided as part of the WNGT 2020 DGT shared task on
``Document-Level Generation and Translation''
\cite{hayashi-etal-2019-findings}.  Statistics on these datasets are
shown in Table~\ref{dataset-stats}.
We used the official train/dev/test splits: 3,398/727/728 for
\textsc{RotoWire}, 22,821/1,739/1,744 for MLB, and 242/240/241 for
German \textsc{RotoWire}. The latter is considerably smaller than its
English counterpart and MLB, and serves to illustrate our model's
sample efficiency when training data is scarce.

All three datasets were preprocessed following the method of
\citet{Puduppully-2020}.  A paragraph plan for an entity is
constructed by verbalizing its records in a fixed sequence of record
type followed by its value.  For example, pitcher \textsl{B.Keller}
from Figure~\ref{fig:example} would be verbalized as
\xml{PLAYER} \cmtt{B.Keller} \xml{H/V} \cmtt{V} \xml{W} \cmtt{7}
\xml{L} \cmtt{5} \xml{IP} \cmtt{8} \xml{PH} \cmtt{4} $\dots$.  We denote
this using the shorthand \xml{V(B.Keller)}.  The paragraph plan for an
event is the verbalization of the players in the event followed by the
verbalization of play-by-plays.  Candidate paragraph plans
$\mathcal{E}$ are obtained by enumerating entities and events and
their combinations (see Table~D in Figure~\ref{fig:example}).  Oracle
macro plans are obtained by matching the mentions of entities and
events in the gold summary with the input table.  We make use of these
oracle macro plans during training.  The versions of MLB and
\textsc{RotoWire} released by \citet{Puduppully-2020} contain
paragraph delimiters for gold summaries; we preprocessed the German
\textsc{RotoWire} in a similar fashion.
Table~\ref{dataset-stats} also shows the average length of the macro
plan in terms of the number of paragraph plans it contains. This is 10.6
for \textsc{RotoWire}, 15.1 for MLB, and 9.5 for German RotoWire.

\paragraph{Training Configuration}
We train our model with the AdaGrad optimizer
\cite{DBLP:journals/jmlr/DuchiHS11} and tune parameters on the
development set. We use a learning rate of 0.15. We learn a joint subword vocabulary
\cite{sennrich-etal-2016-neural} for paragraph plans and summaries
with 6K merge operations for \textsc{RotoWire}, 16K merge operations
for MLB, and 2K merge operations for German \textsc{RotoWire}. The
model is implemented on a fork of OpenNMT-py
\cite{klein-etal-2017-opennmt}.  For efficiency, we batch using
summaries instead of individual paragraphs. %
Batch sizes for MLB, \textsc{RotoWire}, and German-\textsc{RotoWire}
are 8, 5, and 1 respectively.  We set $\lambda$ to 2 in
Equation~\eqref{eq:total_loss}.  In
Equation~\eqref{eq:scheduled_sampling}, $c$ is 1/100000 for MLB,
1/50000 for \textsc{RotoWire}, and 1/30000 for
German-\textsc{RotoWire}. We set the temperature of Gumbel-Softmax to 0.1.

During inference in MLB, similar to \citet{Puduppully-2020}, we block
the repetition of paragraph plan bigrams (i.e.,~we disallow the
repetition of~($z^t, z^{t+1}$)) and select the paragraph plan with the
next higher probability in Equation~\eqref{eq:p_plan}. In addition, we
block consecutive repetitions, and more than two repetitions of a
unigram.  During training we observed high variance in the length of
paragraphs~$y^t$ since the same plan can result in a shorter or longer
paragraph. For example, \xml{V(B.Keller)} corresponds to two
paragraphs (first and third paragraph) with different lengths in
Figure~\ref{fig:example}. We found that this encourages the model to
be conservative and generate relatively short output. We control the
paragraph length \cite{fan-etal-2018-controllable} by creating
discrete bins, each containing approximately an equal number of
paragraphs. During training, we prepend the embedding of the bin to
the current plan~$r^t_z$ (see Equation~\eqref{eq:cross_attn}).  For
inference, bins are tuned on the validation set.

We run inference for 15~paragraphs on \textsc{RotoWire} and German
\textsc{RotoWire}, and for 20~paragraphs on MLB; we stop when the
model predicts the end of paragraph plan token~{\it EOP}.  Unlike
previous work
(\citealp{wiseman-etal-2017-challenges,DBLP:journals/corr/abs-1809-00582,puduppully-etal-2019-data},
\emph{inter alia}), we do not make use of truncated Back Propagation
Through Time (BPTT; \citealp{DBLP:journals/neco/WilliamsP90}), as we
incrementally generate paragraphs instead of long documents.

\paragraph{System Comparisons}
We compared our model with: (1)~a \textbf{Templ}ate-based generator
which creates a document consisting of template sentences. We used
Wiseman et al.'s \shortcite{wiseman-etal-2017-challenges} system on
\textsc{RotoWire} and Puduppully et al.'s
\shortcite{puduppully-etal-2019-data} system on MLB. They are both
similar in that they describe team scores followed by player specific
statistics and a concluding statement.  In MLB, the template
additionally describes play-by-play details.  We also created a
template system for German \textsc{RotoWire} following a similar
approach.  (2)~\textbf{ED$+$CC}, the best performing model of
\citet{wiseman-etal-2017-challenges}. It consists of an
encoder-decoder model equipped with attention and copy mechanisms.
(3)~\textbf{NCP$+$CC}, the micro planning model of
\citet{DBLP:journals/corr/abs-1809-00582}. It first creates a content
plan by pointing to input records through the use of Pointer Networks
\cite{NIPS2015_5866}. The content plan is then encoded with a BiLSTM
and decoded using another LSTM with an attention and copy mechanism.
(4)~\textbf{ENT}, the entity model of
\citet{puduppully-etal-2019-data}.  It creates entity-specific
representations which are updated dynamically. At each time step
during decoding, their model makes use of hierarchical attention by
attending over entity representations and the records corresponding to
these.  (5)~\textbf{MACRO}, the two-stage planning model of
\citet{Puduppully-2020}, which first makes use of Pointer Networks
\cite{NIPS2015_5866} to predict a macro plan from a set of candidate
paragraph plans. The second stage takes the predicted plan as input
and generates the game summary with a sequence-to-sequence model
enhanced with attention and copy mechanisms. In addition, we compare
with a variant of Macro enhanced with length control ($+$Bin).

\section{Results}
\label{sec:results}
Our experiments were designed to explore how the proposed model
compares to related approaches which are either not enhanced with
planning modules or non-incremental. We also investigated the sample
efficiency of these models and the quality of the predicted plans when
these are available. The majority of our results focus on automatic
evaluation metrics. We also follow previous work
\cite{wiseman-etal-2017-challenges,DBLP:journals/corr/abs-1809-00582,
  puduppully-etal-2019-data,Puduppully-2020} in eliciting judgments to
evaluate system output.

\subsection{Automatic Evaluation}

We evaluate model output using BLEU \cite{papineni-etal-2002-bleu}
with the gold summary as a reference. We also report model performance
against the Information Extraction (IE) metrics of
\citet{wiseman-etal-2017-challenges} which are defined based on the
output of an IE model which extracts entity (team and player names)
and value (numbers) pairs from the summary and predicts the type of
relation between them.

Let $\hat{y}$ be the gold summary and $y$ be the model
output. \textsl{Relation Generation} (RG) measures the precision and
count of relations obtained from $y$ that are found in the input
table. \textsl{Content Selection} (CS) measures the precision, recall,
and \mbox{F-measure} of relations extracted from $y$ also found in
$\hat{y}$. And \textsl{Content Ordering} (CO) measures the complement
of the Damerau-Levenshtein distance between relations extracted from
$y$ and $\hat{y}$. Higher values are better for RG Precision, CS
F-measure, CO, and BLEU. We reuse the IE model from
\citet{DBLP:journals/corr/abs-1809-00582} for \textsc{RotoWire},
\citet{Puduppully-2020} for MLB, and
\citet{hayashi-etal-2019-findings} for German \textsc{RotoWire}.  Our
computation of IE metrics for all systems includes duplicate records
\cite{Puduppully-2020}.

In addition to IE-based metrics, we report the number of errors made
by systems according to Number (incorrect number in digits, number
spelled in words, etc.), Name (incorrect names of teams, players, days
of week, etc.), and Word (errors in usage of words) following the
classification of \citet{thomson-reiter-2020-gold}. We detect such
errors automatically using the system of \citet{KasnerTextInContext}
which scored best against gold standard human annotations of the same
type \cite{thomson-reiter-2021-generation}.  We only report these
metrics for English \textsc{RotoWire}, since error annotations (for
automatic metric learning) are not available for other datasets.
Moreover, with regard to Word errors, we only report errors for
incorrect usage of the word \textsl{double-double}.\footnote{A
  double-double occurs when a player scores 10 points or more in two
  record types: points, rebounds, assists, steals, and blocked shots.}
We found such errors to be detected reliably in contrast to Word
errors as a whole for which the precision of the system of
\citet{KasnerTextInContext} is \textasciitilde 50\%.  Lower values are
better for the Number, Name, and double-double errors.  We note metrics
such as RG precision, Number, Name, and double-double errors \emph{directly}
compute the accuracy of the generation model. Metrics
such as CS, CO, and BLEU measure how similar model output is against a
reference summary. Thus, CS, CO and BLEU measure generation accuracy
\emph{indirectly} under the assumption that gold summaries are
accurate.

\begin{table}[t]
\footnotesize
\centering
\begin{tabular}{@{}l@{~}|@{~}c@{~~~}c@{~}|c@{~~~}c@{~~~}c@{~}|@{~}c@{~}|@{~}c@{}} 
\multicolumn{7}{c}{} \\ \thickhline
 \multirow{2}{*}{MLB} &\multicolumn{2}{c|}{RG} &\multicolumn{3}{c@{~}|@{~}}{CS} & CO & \multirow{2}{*}{BLEU}\\

 &\# & P\% & P\% & R\% & F\% & DLD\% & \\ \thickhline
Templ & 62.3 & 99.9 & 21.6 & 55.2 &  31.0 &  11.0 & 4.12 \\ \hline \hline 
ED$+$CC & \textbf{32.5} & 91.3 & 27.8 & 40.6 & 33.0 &   17.1 & 9.68 \\
NCP$+$CC& 19.6 & 81.3 & \textbf{44.5} & 44.1 & 44.3 &  21.9 & 9.68 \\
ENT&23.8 & 81.1 & 40.9 & 49.5 &44.8  &20.7 & 11.50 \\ 
Macro & \dotuline{30.8} & \dotuline{94.4} & 40.8 & \textbf{54.9} & \dotuline{46.8} & \dotuline{21.8} & \dotuline{12.62} \\
~~~$+$Bin & 31.2 & 93.7 & 38.3 & 52.4 & 44.2 & 21.6 & 12.32 \\ \hline \hline 
SeqPlan & 28.9 & \textbf{95.9} & \dotuline{43.3} & \dotuline{53.5} & \textbf{47.8} & \textbf{22.7} & \textbf{14.29} \\
~~~w Uniform & 18.5 & 90.9 & 36.5 & 30.6 & 33.3 & 14.5 & 10.30 \\ 
~~~w Oracle & 27.6 & {95.9} & 42.5 & 50.4 & 46.1 & 22.0 & 13.13 \\
~~~2-Stage& 28.6 & {95.9} & 41.4 & 50.8 & 45.6 & 21.3 & 13.96 \\ \thickhline
\end{tabular}
\vspace*{-.2ex}
\caption{MLB results (test set); relation
  generation (RG) count (\#) and precision (P\%), content selection
  (CS) precision (P\%), recall (R\%), and F-measure (F\%), content
  ordering (CO) as complement of normalized Damerau-Levenshtein
  distance (DLD\%), and BLEU. \textbf{Highest} and \dotuline{second highest} generation models
  are highlighted.} 
\label{tbl:results-with-ie-test-mlb}
\end{table}

\paragraph{MLB Dataset}
Table~\ref{tbl:results-with-ie-test-mlb} summarizes our results on
MLB.  %
Our sequential planning model (SeqPlan) has the highest RG~P among
neural models and performs best in terms of CS~F, CO, and BLEU.  The
variant of Macro with length control ($+$Bin) performs comparably or
worse than Macro.

To examine the importance of latent sequential planning, we also
present a variant of our model which uniformly samples a plan from the
pool~$\mathcal{E}$ instead of Equation~\eqref{eq:p_plan} (see row
w(ith) Uniform in Table~\ref{tbl:results-with-ie-test-mlb}). This
version obtains lower values compared to SeqPlan across all metrics
underscoring the importance of sequential planning.  We also present
two variants of SeqPlan (a)~one which makes use of oracle (instead of
predicted) plans during training to generate~$y^t$; essentially, it
replaces $z^t$ with $z^*$ in Equation~\eqref{eq:generation} (row
w(ith) Oracle in Table~\ref{tbl:results-with-ie-test-mlb}) and (b)~a
two stage model which trains the planner (Equation~\eqref{eq:planner})
and generator (Equation~\eqref{eq:generation}) separately (row 2-stage
in Table~\ref{tbl:results-with-ie-test-mlb}); in this case, we use
greedy decoding to sample $z^t$ from Equation~\eqref{eq:planner}
instead of Gumbel-Softmax and replace~$z^t$ with~$z^*$ in
Equation~\eqref{eq:generation}. Both variants are comparable to
SeqPlan in terms of RG~P but worse in terms of CS~F, CO, and BLEU.

Furthermore, we evaluate the accuracy of the inferred plans by
comparing them against oracle plans, using the CS and CO metrics
(computed over the entities and events in the plan) \footnote{To
  compute the accuracy of macro plans, entities and events from the
  model's plan need to be compared against entities and events in the
  oracle macro plan. \citet{Puduppully-2020} obtained the entities and
  events for the oracle macro plan by extracting these from reference
  summaries.  We noted that this includes coreferent or repeat
  mentions of entities and events within a paragraph.  We instead
  extract entities and events directly from the oracle macro plan.}.
Table~\ref{tbl:inf-plans-cs-co-rw} shows that SeqPlan achieves higher
CS~F and CO scores than Macro.  Again, this indicates planning is
beneficial, particularly when taking the table and the generated
summary into account.

\begin{table}[t]
\footnotesize
\centering
\begin{tabular}{@{}l@{~}|@{~}c@{~~~}c@{~}|c@{~~~}c@{~~~}c@{~}|@{~}c@{~}|@{~}c@{}} 
\multicolumn{7}{c}{} \\ \thickhline
 \multirow{2}{*}{RW} &\multicolumn{2}{c|}{RG} &\multicolumn{3}{c@{~}|@{~}}{CS} & CO & \multirow{2}{*}{BLEU}\\ 
 &\# & P\% & P\% & R\% & F\% & DLD\% & \\ \thickhline

Templ & 54.3 & 99.9 &27.1 &{57.7} & 36.9 &13.1 &8.46  \\ \hline \hline
WS-2017 & 34.1 & 75.1 & 20.3 & 36.3  &26.1 & 12.4 & 14.19 \\
ED$+$CC & 35.9 & 82.6 &19.8 & 33.8 & 24.9 & 12.0 & 14.99 \\ %
NCP$+$CC &{40.8} & {87.6} & 28.0 & {51.1} & 36.2
 &15.8 & \textbf{16.50} \\ 
ENT&32.7 & \dotuline{91.7} & \textbf{34.7} & 48.5 & 40.5 & 16.6 & 16.12 \\
RBF-2020 & 44.9 & 89.5 & 23.9 & 47.0 & 31.7 & 14.3 & 17.16 \\
Macro & 42.1 & \textbf{97.6} & 34.1 & 57.8 &\textbf{42.9} & \textbf{17.7} & 15.46 \\ 
~~~$+$Bin & \textbf{61.0} & 97.2 & 26.8 & \textbf{66.1} & 38.2 & 15.8 & \textbf{16.48} \\ \hline \hline
SeqPlan & \dotuline{46.7} & \textbf{97.6} & \dotuline{30.6} & \dotuline{57.4} & \dotuline{39.9} & \dotuline{16.7} & \dotuline{16.26} \\
~~~w Uniform & 22.0 & 80.2 & 18.2 & 19.6 & 18.9 & 6.0 & 8.61 \\
~~~w Oracle  & 50.4 & 97.2 & 29.0 & 59.1 & 38.9 & 16.8 & 16.32 \\
~~~2-stage & 53.4 & 97.5 & 28.5 & 61.3 & 38.9 & 16.1 & 16.61 \\
\thickhline
\multicolumn{7}{c}{} \\ \thickhline
\multirow{2}{*}{DE-RW} &\multicolumn{2}{c|}{RG} &\multicolumn{3}{c@{~}|@{~}}{CS} & CO & \multirow{2}{*}{BLEU}\\

 &\# & P\% & P\% & R\% & F\% & DLD\% & \\ \thickhline
Templ & 54.4 & 99.9 & 17.2 & 63.0 &  27.1 &  11.6 & 7.32 \\ \hline \hline
ED$+$CC& \dotuline{24.8} & {59.3} & 6.7 & 18.8 & 9.9 & 6.8 & 5.09 \\
NCP$+$CC& 17.7 & 52.5 & 11.3 & \dotuline{25.7} & 15.7 & 9.6 & \dotuline{7.29} \\
ENT& 17.4 & \dotuline{64.7} & \dotuline{13.3} & 24.0 & \dotuline{17.1} & \dotuline{9.8} & 6.52 \\
RBF-2020 & 0.2 & 4.0 & 1.1 & 0.4 & 0.6 & 0.3 &2.29 \\
Macro & \textbf{30.2} & 49.7 & 5.1 & 21.0 & 8.3 & 6.1 & 5.15 \\
~~~$+$Bin & 20.4 & 55.0 & 7.9 & 20.0 & 11.3 & 8.1 & 6.18 \\
\hline\hline     
SeqPlan & 13.8 & \textbf{91.8} & \textbf{38.0} & \textbf{38.4} & \textbf{38.2} & \textbf{21.2} & \textbf{8.65} \\
\thickhline

\end{tabular}
\vspace*{-.2ex}
\caption{Evaluation on
  \textsc{RotoWire} (RW) and German \textsc{RotoWire} (DE-RW) test sets; relation
  generation (RG) count (\#) and precision (P\%), content selection
  (CS) precision (P\%), recall (R\%), and F-measure (F\%), content
  ordering (CO) as complement of normalized Damerau-Levenshtein distance (DLD\%),
  and BLEU.  \textbf{Highest} and \dotuline{second highest} generation models
  are highlighted.}
\label{tbl:results-with-ie-test} 
\end{table}

\paragraph{English and German \textsc{RotoWire}}
Results on \textsc{RotoWire} are presented in
Table~\ref{tbl:results-with-ie-test} (top). In addition to Templ,
ED$+$CC, NCP$+$CC, and ENT, we compare with the models of
\citet{wiseman-etal-2017-challenges} (WS-2017) and
\citet{rebuffel2020hierarchical} (RBF-2020). WS-2017 is the best
performing model of \citet{wiseman-etal-2017-challenges}. Note that
ED$+$CC is an improved re-implementation of WS-2017. RBF-2020
represents the current state-of-the-art on \textsc{RotoWire}, and
comprises of a Transformer encoder-decoder architecture
\cite{NIPS2017_7181} with hierarchical attention on entities and their
records. The models of \citet{saleh-etal-2019-naver},
\citet{iso-etal-2019-learning}, and \citet{gong-etal-2019-table} are
not comparable as they make use of information additional to the table
such as previous/next games or the author of the game summary. The
model of \citet{narayan-etal-2020-stepwise} is also not comparable as
it relies on a pretrained language model
\cite{rothe-etal-2020-leveraging} to generate the summary sentences.

Table~\ref{tbl:results-with-ie-test} (bottom) shows our results on
German \textsc{RotoWire}. We compare against NCP$+$CC's entry in the
WNGT 2019 shared task\footnote{We thank Hiroaki Hayashi for providing
  us with the output of the NCP$+$CC system.}
\cite{hayashi-etal-2019-findings}, and our implementation of Templ, ED$+$CC, ENT, Macro
and RBF-2020.
\citet{saleh-etal-2019-naver} are not comparable as they pretrain on
32M parallel and 420M monolingual data. Likewise,
\citet{puduppully-etal-2019-university} make use of a jointly trained
multilingual model by combining \textsc{RotoWire} with German
\textsc{RotoWire}.

We find that SeqPlan achieves highest RG~P amongst neural models, and
performs on par with Macro (it obtains higher BLEU but lower CS~F and
CO scores).  The $+$Bin variant of Macro performs better on BLEU but
worse on other metrics.  As in
Table~\ref{tbl:results-with-ie-test-mlb}, w~Uniform struggles across
metrics corroborating our hypothesis that latent sequential planning
improves generation performance.  The other two variants (w~Oracle and
2-Stage) are worse than SeqPlan in RG~P and CS~F, comparable in
CO, and slightly higher in terms of BLEU.

On German, our model is best across metrics achieving an RG~P
of~91.8\% which is higher by 42\% (absolute) compared to of Macro. In
fact, the RG~P of SeqPlan is superior to
\citet{saleh-etal-2019-naver} whose model is pretrained with
additional data and is considered state of the art
\cite{hayashi-etal-2019-findings}.  RG\# is lower mainly because of a
bug in the German IE which excludes number records. RG\# for NCP$+$CC
and Macro is too high because the summaries contain a lot of
repetition.  The same record will repeat at least once with NCP$+$CC
and three times with Macro, whereas only~7\% of the records are
repeated with SeqPlan. %

Table~\ref{tbl:inf-plans-cs-co-rw} evaluates the quality of the plans
inferred by our model on the \textsc{RotoWire} dataset. As can be
seen, SeqPlan is slightly worse than Macro in terms of CS~F and CO. We
believe this is because summaries in \textsc{RotoWire} are somewhat
formulaic, with a plan similar to Templ: an opening statement is
followed by a description of the top scoring players, and a conclusion
describing the next match. Such plans can be learnt well by Macro
without access to the summary. MLB texts show a lot more diversity in
terms of length, and the sequencing of entities and events. The
learning problem is also more challenging, supported by the fact that
the template system does not do very well in this domain (i.e., it is
worse in BLEU, CS~F, and CO compared to \textsc{RotoWire}).  In German
\textsc{RotoWire}, SeqPlan plans achieve higher CS~F and CO than
Macro.

\begin{table}[t]
\footnotesize
\centering
\begin{tabular}{ll|lcc|c} 
\multicolumn{5}{c}{} \\ \thickhline
 \multicolumn{2}{c}{}  &\multicolumn{3}{c}{CS} & CO \\ 
 \multicolumn{2}{c|}{\raisebox{.4ex}[0pt]{Datasets}} &  P\% & R\% & F\% & DLD\% \\ \thickhline
&Macro & 73.6 & 45.9 & 56.5 & 27.0 \\
\raisebox{-0.5ex}[0pt]{\begin{sideways}MLB~~\end{sideways}} &SeqPlan & 74.4 & 51.1 & 60.6 & 27.1 \\\hline\hline
&Macro & 81.5 & 62.7 & 70.9 & 36.3 \\  %
\raisebox{.1ex}[0pt]{\begin{sideways}RW\end{sideways}}&SeqPlan & 79.1
& 61.6 & 69.3& 35.5 \\\hline \hline
 & & & & & \\
&\raisebox{2ex}[0pt]{Macro} & \raisebox{2ex}[0pt]{86.8} & \raisebox{2ex}[0pt]{34.2} & \raisebox{2ex}[0pt]{49.0} & \raisebox{2ex}[0pt]{30.1} \\
\raisebox{.02ex}[0pt]{\begin{sideways}DE-RW\end{sideways}} &\raisebox{1.5ex}[0pt]{SeqPlan} & \raisebox{1.5ex}[0pt]{73.1} & \raisebox{1.5ex}[0pt]{60.8} & \raisebox{1.5ex}[0pt]{66.4} & \raisebox{1.5ex}[0pt]{31.0} \\
\thickhline 
\end{tabular}
\caption{Evaluation of macro planning stage (test set); 
content selection (CS) precision (P\%), recall (R\%), and F-measure
(F\%), content ordering (CO) as  complement of normalized
Damerau-Levenshtein distance (DLD\%).}
\label{tbl:inf-plans-cs-co-rw}
\end{table}

\begin{table}
\centering
\footnotesize
\begin{tabular}{ l|c|c|c} 
 \thickhline
 & Number & Name & double-double\\
 \thickhline
 Templ & \hspace*{1ex}0.08*&3.05* & 0.00*\\ \hline \hline
WS-2017 & 13.01*& 9.66* & 0.36*\\
 ED$+$CC & \hspace*{1ex}8.11*& 8.29* & 0.31*\\ 
 NCP$+$CC&\hspace*{1ex}7.89* & 7.76* & \hspace*{-1ex}0.14\\
 ENT & \hspace*{1ex}5.89* &7.24* & \hspace*{-1ex}0.15\\
RBF-2020 &\hspace*{1ex}6.20* &8.39*& 0.41*\\
Macro & 2.57 &4.60*  & \hspace*{-1ex}0.18\\
SeqPlan & 2.70& \hspace*{-1ex}6.56 & \hspace*{-1ex}0.20\\
 \thickhline
\end{tabular}
\caption{Number, Name, and double-double (Word) errors per
  example. Systems significantly different from   SeqPlan are marked
  with an asterisk~* (using a one-way ANOVA with
  posthoc Tukey HSD tests; \mbox{$p\leq0.05$}).}
\label{tab:fact-eval}
\end{table}

Table~\ref{tab:fact-eval} reports complementary automatic metrics on
English \textsc{RotoWire} aiming to assess the factuality of generated
output.  We find that Templ has the least Number, Name, and
double-double errors. This is expected as it simply reproduces facts
from the table.  SeqPlan and Macro have similar Number errors, and
both are significantly better than other neural models. SeqPlan has
significantly more Name errors than Macro, and significantly fewer
than other neural models.  Inspection of Name errors revealed that
these are mostly due to incorrect information about next games. Such
information is not part of the input and models are prone to
hallucinate.  SeqPlan fares worse as it attempts to discuss next games
for both teams while Macro focuses on one team only. In terms of
double-double errors, SeqPlan is comparable to Macro, ENT and
NCP$+$CC, and significantly better than WS-2017, ED$+$CC, and
RBF-2020.

\begin{figure}[t]
\begin{tabular}{@{\hspace*{-.25cm}}c@{}c@{}}
 \includegraphics[width=0.26\textwidth]{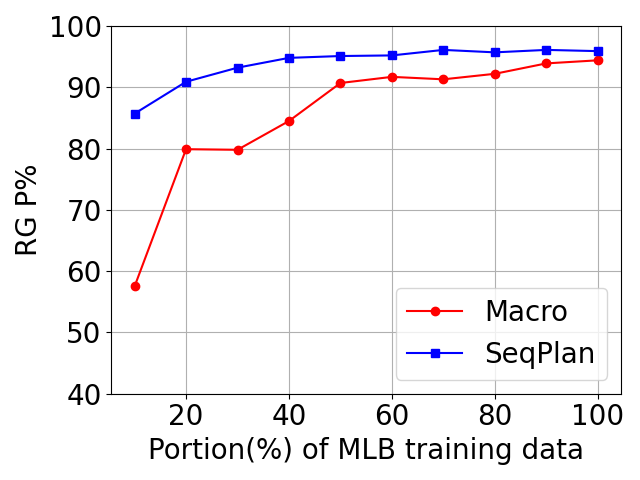} &
\includegraphics[width=0.26\textwidth]{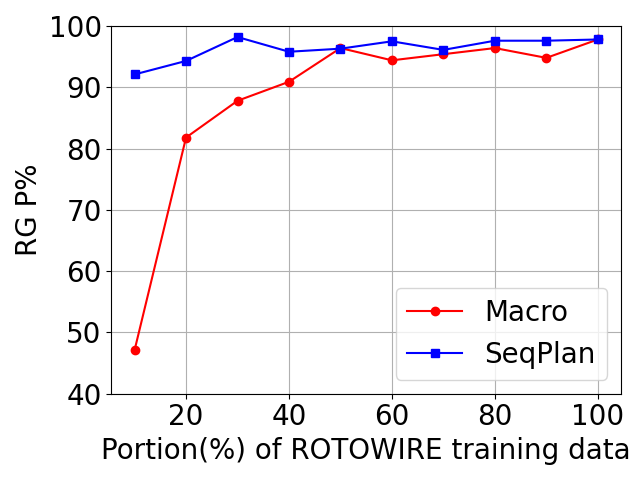}\\
\small{(a)} & \small{(b)}\\
\end{tabular}
\caption{Sample efficiency for (a)~MLB and (b)~\textsc{RotoWire}
  datasets. SeqPlan and Macro are trained on different portions (\%) of the
  training dataset and performance is measured with RG P\%.}
\label{fig:sample-efficiency-mlb}
\end{figure}

\subsection{Sample Efficiency}

We also evaluated whether SeqPlan is more sample efficient in
comparison to Macro, by examining how RG~P varies with (training)
data size. As shown in Figure~\ref{fig:sample-efficiency-mlb}, the
difference between SeqPlan and Macro is more pronounced when
relatively little data is available. For example, with 10\% of
training data, RG~P for SeqPlan on MLB is~85.7\% and 92.1\% on
\textsc{RotoWire}.  In contrast, Macro obtains~57.5\% on MLB and
47.1\% on \textsc{RotoWire}.  As more training data becomes available,
the difference in RG~P decreases.  The slope of increase in RG~P
for Macro is higher for \textsc{RotoWire} than MLB. We hypothesize
this is because MLB has longer summaries with more paragraphs, and is
thus more difficult for Macro to learn alignments between paragraph
plans and text paragraphs in the game summary.

\begin{table}[t]
\footnotesize
\centering
\begin{tabular}{@{}l@{~~}r@{~~}r@{~~}c@{~~}r@{~~}c@{}}
\thickhline
MLB & \#Supp & {\#Contra} & Gram & Coher & Concis \\ \thickhline
Gold    & 3.59~~ & 0.14~~ & \hspace*{1.6ex}21.67~ & 29.17~~ &\hspace*{1.1ex}14.17\\
Templ   & 4.21* & 0.04~~ & $-$58.33* &  $-$48.33*&\hspace*{1.7ex}9.17\\
ED$+$CC & 3.42~~ & 0.72* & $-$32.50* & $-$18.33* &$-$48.33*\\
Macro  & 3.76~~ & 0.25~~ & \hspace*{1.2ex}37.50~ & 15.00~~~&\hspace*{.8ex}22.50\\
SeqPlan & 3.68~~ & 0.19~~ & \hspace*{1.2ex}31.67~ & 22.50~~~&\hspace*{1.5ex}2.50 \\
\thickhline
\multicolumn{6}{c}{} \\
\thickhline
\textsc{RotoWire} & \#Supp & \#Contra & Gram & Coher & Concis \\
\thickhline
Gold    & 3.63* & 0.07~~ & \hspace*{2ex}42.67*& 40.67~~~& \hspace*{.5ex}28.00\\
Templ   & 7.57* & 0.08~~ & $-$57.33* &$-$55.33*& $-$34.67*\\
ED$+$CC  & 3.92~~ & 0.91* &\hspace*{2ex}4.00 &$-$14.67*&\hspace*{-1ex}$-$13.33\\
RBF-2020   & 5.08~~ & 0.67* & \hspace*{2ex}6.00&1.33~~~&\hspace*{-.1ex}$-$0.67\\
Macro  & 4.00~~ & 0.27~~ & \hspace*{2ex}0.67&7.33~~~&10.00\\ 
SeqPlan &4.84~~ & 0.17~~ & \hspace*{2ex}4.00&20.67~~~&10.67\\
\thickhline

\end{tabular}
\caption{Average number of supported (\#Supp) and contradicting
  (\#Contra) facts in game summaries and \textit{best-worst scaling}
  evaluation for Coherence (Coher), Conciseness (Concis), and Grammaticality
  (Gram). Lower is better for contradicting facts; higher is better for Coherence, Conciseness, and Grammaticality.  
  Systems significantly different from
  SeqPlan are marked with an asterisk * (using a one-way ANOVA with
  posthoc Tukey HSD tests; \mbox{$p\leq0.05$}).  }.
  \label{tbl:mlb-human-eval}
\end{table}

\subsection{Human Evaluation}
We used the Amazon Mechanical Turk (AMT) crowdsourcing platform for
our judgment elicitation study. To ensure consistent ratings
\cite{van-der-lee-etal-2019-best} we required that raters have
completed at least 1,000 tasks, and have at least 98\% approval
rate. Participants were restricted to English speaking countries (USA,
UK, Canada, Australia, Ireland, or New Zealand) and were allowed to
provide feedback or ask questions. Raters were paid an average of
0.35\$ for each task, ensuring that the remuneration is higher than
the minimum wage per hour in the US.  We compared SeqPlan with Gold,
Templ, ED$+$CC, and Macro; we did not compare against ENT as previous
work \cite{Puduppully-2020} has shown that it performs poorly against
Macro.  For \textsc{RotoWire}, we additionally compared against
RBF-2020.

\paragraph{Supported and Contradicted Facts}
Our first eliciation study provided raters with box scores (and
play-by-plays in the case of MLB), along with sentences randomly
extracted from game summaries. We asked them to count supported and
contradicting facts (ignoring hallucinations).
Participants were given a cheatsheet to help them understand box score
and play-by-play statistics as well as examples of sentences with the
correct count of supported and contradicting facts.  This evaluation
was conducted on 40 summaries (20 for each dataset), with four
sentences per summary, each rated by three participants.  For MLB,
this resulted in 300 tasks (5 systems $\times$ 20 summaries $\times$ 3
raters) and for \textsc{RotoWire} in 360 (6 systems $\times$ 20
summaries $\times$ 3 raters). Altogether, we had 177 participants.
The agreement between raters using Krippendorff's $\alpha$ for
supported facts and contradicting facts was~0.43.

Table~\ref{tbl:mlb-human-eval} (columns \#Supp and \#Contra) presents
our results. Lower is better for contradicting facts. In case of
supporting facts, the count should neither be too high nor too low. A
high count of supporting facts indicates indicates poor content
selection. A low count of supporting facts with a high count of
contradicting facts indicates low accuracy of generation.

Templ achieves the lowest count of contradicting facts and the
highest count of supported facts for both the datasets. This is
no surprise as it essentially regurgitates facts (i.e., records) from
the table.
On MLB, all systems display a comparable count of
\emph{supported} facts (differences are not statistically
significant), with the exception of Templ which contains significantly
more. In terms of \emph{contradicting} facts, SeqPlan performs on par
with Macro, Gold and Templ, and is significantly better than ED$+$CC.
On \textsc{RotoWire}, %
in terms of supported facts, SeqPlan performs on par with the other neural models, 
is significantly higher than Gold, and
significantly lower than Templ.
In terms of contradicting facts, SeqPlan performs on par with
Macro, Gold and Templ, and significantly better than
ED$+$CC and RBF-2020.

\paragraph{Coherence, Grammaticality, and Conciseness}
In our second study, raters were asked to choose the better summary
from a pair of summaries based on \emph{Coherence} (is the summary
well structured and well organized and does it have a natural ordering
of the facts?), \emph{Conciseness} (does the summary avoid unnecessary
repetition including whole sentences, facts or phrases?), and
\emph{Grammaticality} (is the summary written in well-formed
English?).  For this study, we required that the raters be able to
comfortably comprehend summaries of NBA/MLB games.  We obtained
ratings using Best-Worst scaling
\cite{louviere1991best,louviere2015best}, an elicitation paradigm
shown to be more accurate than Likert scales.  The score for a system
is obtained by the number of times it is rated best minus the number
of times it is rated worst \cite{orme2009maxdiff}.  Scores range
between $-$100 (absolutely worst) and $+$100 (absolutely best); higher is better.  We
assessed 40 summaries from the test set (20 for each dataset).  Each
summary pair was rated by three participants. For MLB, we created
1,800 tasks (10 system pairs $\times$ 20 summaries $\times$ 3 raters
$\times$ 3~dimensions) and 2,700 for \textsc{RotoWire} (15 pairs of
systems $\times$ 20 summaries $\times$ 3 raters $\times$ 3
dimensions). Altogether, 377 raters participated in this task.  The
agreement between the raters using Krippendorff's~$\alpha$ was~0.49.

On MLB, SeqPlan is significantly more {coherent} than 
ED$+$CC and Templ, and is comparable with Gold and Macro. A
similar picture emerges with {grammaticality}.
SeqPlan is as
concise as Gold, Macro and Templ, and significantly better than
ED$+$CC. 
On \textsc{RotoWire},
SeqPlan is significantly more coherent than Templ and ED$+$CC,
but on par with Macro, RBF-2020 and Gold. In terms of conciseness, SeqPlan is comparable with Gold,
Macro, RBF-2020, and ED+CC, and significantly better than Templ. 
In terms of grammaticality, SeqPlan is comparable with
Macro, RBF-2020, and ED+CC, significantly better than Templ,
and significantly worse than Gold.

\begin{table}[t]
\scriptsize
\centering
\begin{tabular}{@{}p{7.7cm}@{}} 
  \thickhline
  \xml{V(Cardinals)} $\rightarrow$ 
  \xml{V(9-T)} $\rightarrow$
  \xml{V(Cardinals)} $\rightarrow$ 
  \xml{V(Cardinals)} \xml{V(Brewers)}$\rightarrow$ 
  \xml{V(9-T)} $\rightarrow$
  \xml{V(8-B)} $\rightarrow$
  \xml{V(8-T)} $\rightarrow$
  \xml{V(8-B)} $\rightarrow$
  \xml{V(9-B)} $\rightarrow$
  \xml{V(Brewers)}$\rightarrow$ 
  \xml{V(Adam Wainwright)}$\rightarrow$ 
  \xml{V(Brewers)}$\rightarrow$ 
  \xml{V(3-T)} $\rightarrow$
  \xml{V(3-B)} $\rightarrow$
  \xml{V(Carlos Villanueva)}
  \\ \hline 
  ST. LOUIS -- The \textbf{St. Louis Cardinals} have been waiting for
  their starting rotation. \textcolor{blue}{\xml{P}} Skip Schumaker
  drove in the go-ahead run with a double in the \textbf{ninth}
  inning, and the Cardinals beat the Milwaukee Brewers 4--3 on
  Wednesday night to avoid a three-game
  sweep. \textcolor{blue}{\xml{P}} The \textbf{Cardinals} have won
  four of five, and have won four in a row. \textcolor{blue}{\xml{P}}
  The \textbf{Cardinals} have won four of five, including a three-game
  sweep by the \textbf{Brewers}. \textcolor{blue}{\xml{P}} Brian
  Barton led off the \textbf{ninth} with a pinch-hit double off
  Derrick Turnbow (0--1) and moved to third on Cesar Izturis' sacrifice
  bunt. Schumaker drove in Barton with a double down the left-field
  line. \textcolor{blue}{\xml{P}} Ryan Braun, who had two hits, led
  off the \textbf{eighth} with a double off Ryan Franklin (1--1). Braun
  went to third on a wild pitch and scored on Corey Hart's triple into
  the right-field corner. \textcolor{blue}{\xml{P}} Albert Pujols was
  intentionally walked to load the bases with one out in the
  \textbf{eighth}, and Guillermo Ankiel flied out. Troy Glaus walked
  to load the bases for Kennedy, who hit a sacrifice fly off Guillermo
  Mota. \textcolor{blue}{\xml{P}} Ryan Franklin (1--1) got the win
  despite giving up a run in the \textbf{eighth}. Ryan Braun led off
  with a double and scored on Corey Hart's one-out
  triple. \textcolor{blue}{\xml{P}} Jason Isringhausen pitched a
  perfect \textbf{ninth} for his seventh save in nine chances. He has
  converted his last six save opportunities and has n't allowed a run
  in his last three appearances. \textcolor{blue}{\xml{P}} The
  \textbf{Brewers} lost for the seventh time in eight
  games. \textcolor{blue}{\xml{P}} \textbf{Wainwright} allowed two
  runs and four hits in seven innings. He walked four and struck out
  six. \textcolor{blue}{\xml{P}} \textbf{Brewers} manager Ron Roenicke
  was ejected by home plate umpire Bill Miller for arguing a called
  third strike. \textcolor{blue}{\xml{P}} The Cardinals took a 2--0
  lead in the \textbf{third}. Albert Pujols walked with two outs and
  Rick Ankiel walked. Glaus then lined a two-run double into the
  left-field corner. \textcolor{blue}{\xml{P}} The Brewers tied it in
  the \textbf{third}. Jason Kendall led off with a double and scored
  on Rickie Weeks' double. Ryan Braun's RBI single tied it at
  2. \textcolor{blue}{\xml{P}} \textbf{Villanueva} allowed two runs
  and three hits in seven innings. He walked four and struck out one. 
  \\

\thickhline
\end{tabular}
\vspace*{-1.5ex}
\caption{Predicted macro plan (top) and generated output from our
  model. Transitions between paragraph plans are shown using
  $\rightarrow$. Paragraphs are separated with \textcolor{blue}{\xml{P}}
  delimiters. Entities and events in the summary corresponding to the
  macro plan are boldfaced.  }
\label{tab:output}
\end{table}

\section{Discussion}

In this work, we proposed a novel sequential latent variable model for
joint macro planning and generation. Key in our approach is the
creation of a latent plan in a sequential manner, while interleaving
the prediction of plans and the generation of corresponding
paragraphs.  We proposed to deconstruct monolithic long document
generation into smaller units (paragraphs in our case) which affords
flexibility and better communication between planning and generation.
Taken together, the results of automatic and human evaluation suggest
that SeqPlan performs best in terms of factuality and coherence, it
generates diverse, and overall fluent summaries and is less
data-hungry compared to strong systems like Macro and NCP$+$CC.  As
SeqPlan does not have to learn alignments between the macro plan and
the output text, it is better suited for long-form
generation. Potential applications include summarizing books
\cite{kryscinski2021booksum} where the output can be longer than 1,000
tokens or generating financial reports
\cite{kogan-etal-2009-predicting, handschke-etal-2018-corpus} where
the output exceeds 9,000 tokens. Existing approaches for long-form
generation summarize individual paragraphs independently
\cite{kryscinski2021booksum} or adopt a hierarchical approach
\cite{DBLP:journals/corr/abs-2109-10862} where summaries of paragraphs
form the basis of chapter summaries which in turn are composed into a
book summary.

Table~\ref{tab:output} gives an example of SeqPlan output. We see that
the game summary follows the macro plan closely. In addition, the
paragraph plans and the paragraphs exhibit coherent ordering.  Manual
inspection of SeqPlan summaries reveals that a major source of errors
in MLB relate to attention diffusing over long paragraph plans.  As an
example, consider the following paragraph produced by SeqPlan
``\textsl{Casey Kotchman had three hits and three RBIs , including a
  two-run double in the second inning that put the Angels up
  2--0. Torii Hunter had \textcolor{red}{three} hits and drove in a
  run~.}''  In reality, \textsl{Torii Hunter} had two hits but the
model incorrectly generates hits for \textsl{Casey Kotchman}.  The
corresponding paragraph plan is 360 tokens long and attention fails to
discern important tokens.  A more sophisticated encoder, e.g., based
on Transformers \cite{NIPS2017_7181}, could make attention more
focused. In \textsc{RotoWire}, the majority of errors involve numbers
(e.g., team attributes) and numerical comparisons.  Incorporating
pre-executed operations such as min, max
\cite{nie-etal-2018-operation} could help alleviate these
errors.  %

Finally, it is worth mentioning that although the template models
achieve highest RG precision for both MLB and \textsc{RotoWire}
(Tables \ref{tbl:results-with-ie-test-mlb} and~\ref{tbl:results-with-ie-test}), this is mainly because they repeat
facts from the table. Template models score low against CS~F, CO, and
BLEU metrics. In addition, they obtain lowest scores in Grammaticality
and Coherence (Table~\ref{tbl:mlb-human-eval}) which indicates that
they are poor at selecting records from the table and ordering them
correctly in a fluent manner.

\section*{Acknowledgements}
We thank the Action Editor, Ehud Reiter, and the anonymous reviewers 
for their constructive feedback. 
We also thank Parag Jain for
helpful discussions. 
We acknowledge the financial support of the European Research Council 
(award number 681760, ``Translating Multiple Modalities into Text'').

\bibliography{anthology,custom}
\bibliographystyle{acl_natbib}

\end{document}